\documentclass{article}
\usepackage{spconf,amsmath,graphicx}
\usepackage{amssymb,bm}
\usepackage{float}
\usepackage{colortbl}
\usepackage{subfigure}
\usepackage[english]{babel}
\usepackage{amsthm}
\usepackage[dvipsnames]{xcolor}
\definecolor{mygreen}{RGB}{0, 199, 0}
\definecolor{myorange}{RGB}{250, 100, 0}
\definecolor{myred}{RGB}{200, 0, 0}
\definecolor{myblue}{RGB}{30, 144, 255}
\definecolor{mylightskyblue}{RGB}{135, 206, 250}
\definecolor{myskyblue}{RGB}{0, 191, 255}
\definecolor{mypowderblue}{RGB}{176, 196, 222}
\newtheorem{theorem}{Theorem}[section]
\newtheorem{proposition}[theorem]{Proposition}
\newtheorem{remark}[theorem]{Remark}

\title{Robust classification with Flexible Discriminant Analysis in heterogeneous data}
\twoauthors{P. Houdouin, F. Pascal\sthanks{The work of F. Pascal has been partially supported by DGA under grant ANR-17-ASTR-0015.}, M. Jonckheere}
{Universit\'e Paris-Saclay, CNRS, CentraleSup\'elec,\\ 
Laboratoire des signaux et syst\`emes \\
91190, Gif-sur-Yvette, France \\
\{pierre.houdouin, frederic.pascal\}@centralesupelec.fr
}
{A. Wang}
{Department of Engineering \\
University of Cambridge \\
aslw3@cam.ac.uk \\
https://github.com/Andrewwango/femda}

\begin{document}
%
\maketitle
\begin{abstract}
Linear and Quadratic Discriminant Analysis are well-known classical methods but can heavily suffer from non-Gaussian distributions and/or contaminated datasets, mainly because of the underlying Gaussian assumption that is not robust. To fill this gap, this paper presents a new robust discriminant analysis where each data point is drawn by its own arbitrary Elliptically Symmetrical (ES) distribution and its own arbitrary scale parameter. Such a model allows for possibly very heterogeneous, independent but non-identically distributed samples. After deriving a new decision rule, it is shown that maximum-likelihood parameter estimation and classification are very simple, fast and robust compared to state-of-the-art methods.
\end{abstract}
\section{Introduction}
\label{sec:intro}

Discriminant analysis is a widely used statistical tool to perform classification tasks. Historical discriminant analysis \cite{2} assumes that observations are drawn from Gaussian distributions and the decision rule consists in choosing the cluster that maximizes the likelihood of the observation. However, when the underlying assumption fails to hold, the impact on the result can be significant. In early 80s, \cite{3} and \cite{4} studied the impact of contamination and mislabelling on the performances of such methods and \cite{5} how non-normality impacts Quadratic Discriminant Analysis (QDA). To tackle such sensitivity, \cite{8} suggests the use of robust M-estimators. The major drawback is its low breakdown point in high dimensions, but \cite{12} came up with a robust S-estimator to alleviate this issue.

More recently, \cite{20} dropped the Gaussian distribution hypothesis for the underlying distributions to replace it by the more general case of multivariate $t$-distribution. In 2015, \cite{21} generalized discriminant analysis methods to elliptical symmetric (ES) distributions (see \cite{ollila2012complex} for a review on these distributions). The new method called Generalized QDA (GQDA) relies on the estimation of a threshold parameter, whose optimal value is fixed for each sub-family of distribution. The case $c=1$ corresponds to the Gaussian case. Finally, \cite{22} improved the previous work by adding robust estimators, coming up with the Robust GQDA (RGQDA) method. 

All these methods assume that all clusters belong to the same distribution family. In practice, such an hypothesis may not hold. Inspired by \cite{roizman2019flexible}, this paper proposes a new method that does not assume any prior on the underlying distributions, and allows for each observation to be drawn from a different family of distribution. Points in the same cluster do not need to be identically distributed, only to be drawn independently. The counterpart to such flexibility relies in the characteristics of the clusters. Indeed, we assume that points in the same cluster are drawn from distributions that share the same mean and scatter matrix. However, under assumption of existence, points in the same cluster have only proportional covariances matrices. 

The paper is organized as follows. Section \ref{sec:2} presents the model and provides the main derivation of this work.Then, Section \ref{sec:3} contains simulations on synthetic data while experiments on real data are presented in Section \ref{sec:4}. Concluding remarks and perspectives are drawn in Section \ref{sec:5}.

\section{Flexible EM-inspired discriminant analysis}
\label{sec:2}

\indent\textbf{Statistical model:} Let us assume that each observation $\mathbf{x}_i$ is drawn from an ES distribution. The mean and scatter matrix depend on the cluster to which the point belongs to while the nuisance parameter $\tau_{i,k}$ may depend on both observation $i$ and class $k$. We then have the following probability density function for $\mathbf{x}_i \in \mathcal{C}_k$ : 
$$
f(\mathbf{x}_i) =  A_{i} \left| \mathbf{\Sigma}_k \right|^{-\frac{1}{2}} \tau_{i,k}^{-\frac{m}{2}} g_{i} \left( \frac{(\mathbf{x}_i-\boldsymbol{\mu}_k)^T \mathbf{\Sigma}_k^{-1} (\mathbf{x}_i-\boldsymbol{\mu}_k)}{\tau_{i,k}} \right)
$$
\textbf{Expression of the log-likelihood and Maximum Likelihood estimators:} Given $n_k$ independent observations $\mathbf{x}_1,...,\mathbf{x}_{n_k}$ in class $\mathcal{C}_k$, the log-likelihood of the sample can be rewritten as follows: 
\begin{equation}\label{log-likelihood}
l(\mathbf{x}_1,...,\mathbf{x}_{n_k})
    = \sum_{i=1}^{n_k} \log \left( A_i \left| \mathbf{\Sigma}_k \right|^{-\frac{1}{2}} t_{i,k}^{-\frac{m}{2}}\, s_{i,k}^{\frac{m}{2}}\, g_i(s_{i,k}) \right) 
\end{equation}
where $t_{i,k} = (\mathbf{x}_i-\boldsymbol{\mu}_k)^T \mathbf{\Sigma}_k^{-1}(\mathbf{x}_i-\boldsymbol{\mu}_k)$ and $s_{i,k}=t_{i,k}/\tau_{i,k}$. 
Then, maximizing Eq. \eqref{log-likelihood} w.r.t. $\tau_{i,k}$, for fixed $\boldsymbol{\mu}_k$ and $\mathbf{\Sigma}_k$ leads to
$$
\hat{\tau}_{i,k} = \frac{(\mathbf{x}_i-\boldsymbol{\mu}_k)^T \mathbf{\Sigma}_k^{-1} (\mathbf{x}_i-\boldsymbol{\mu}_k)}{\arg \max_{t \in \mathbb{R}^+} \{t^{\frac{m}{2}} g_i(t) \}}.
$$ 
Due to the assumptions on $g_i$, the denominator always exists. Replacing $\tau_{i,k}$ by $\hat{\tau}_{i,k}$ in Eq.\eqref{log-likelihood} leads to:

\begin{equation*}
l(\mathbf x_i) = \tilde{A_i} - \frac{1}{2} \log \left(\left| \mathbf{\Sigma}_k \right| \left( (\mathbf{x}_i-\boldsymbol{\mu}_k)^T \mathbf{\Sigma}_k^{-1}(\mathbf{x}_i-\boldsymbol{\mu}_k) \right)^{m}\right)   
\end{equation*}
where $\tilde{A_i} = \log(A_i) + \log(\max_{t \in \mathbb{R}^+} \{t^{\frac{m}{2}} g_i(t) \}).$

At this stage, one can notice that the flexibility in the choice of the covariance matrix scale allows us to make the impact of the generator function in the likelihood boil down to a multiplicative constant that does not depend on $k$. One obtains robust estimators (derived however using MLE) for the mean and scatter matrix as follows: 

\begin{equation}
\left\{\begin{array}{ccl}
\hat{\boldsymbol{\mu}}_k & = & \displaystyle \cfrac{\sum_{i=1}^{n_k} w_{i,k} \mathbf{x}_i}{\sum_{i=1}^{n_k} w_{i,k}},  \\
\hat{\mathbf{\Sigma}}_k & = & \displaystyle\frac{m}{n_k} \sum_{i=1}^{n_k} w_{i,k} (\mathbf{x}_i-\hat{\boldsymbol{\mu}}_k)(\mathbf{x}_i-\hat{\boldsymbol{\mu}}_k)^T
\end{array}
\right.
\end{equation}
where $w_{i,k} = 1/t_{i,k}$.

Note that $\hat{\boldsymbol{\mu}}_k$ is insensitive to the scale of $\hat{\mathbf{\Sigma}}_k$ and if $\hat{\mathbf{\Sigma}}_k$ is a solution to the fixed-point equation, $\lambda \hat{\mathbf{\Sigma}}_k$ is also solution. Estimators obtained are close to robust M-estimators, but with weights proportional to the squared Mahalanobis distance. The convergence of the two fixed-point equations has been analyzed in  \cite{roizman2019flexible}.\\

\noindent\textbf{Classification rule:} Equipped with these estimators, used for the training part of discriminant analysis, one can now derive one of the contribution of this work, the classification rule. this is the following proposition.

\begin{proposition}
The decision rule for Flexible EM-inspired discriminant analysis (FEMDA) is given by
\begin{equation}\label{femda_rule}
\mathbf x_i \in \mathcal{C}_k \iff \left(\forall j \neq k, \Delta_{jk}^2(\mathbf x_i) \geq \frac{1}{m} \lambda_{jk} \right)
\end{equation}
with $\Delta_{jk}^2(\mathbf x_i) = \log \left(\cfrac{(\mathbf{x}_i- \boldsymbol{\mu}_j)^T \mathbf{\Sigma}_j^{-1}(\mathbf{x}_i- \boldsymbol{\mu}_j)}{(\mathbf{x}_i-\boldsymbol{\mu}_k)^T \mathbf{\Sigma}_k^{-1}(\mathbf{x}_i-\boldsymbol{\mu}_k)}\right)$ and $\lambda_{ik} = \log \left(\cfrac{\left| \mathbf{\Sigma}_k \right|}{\left| \mathbf{\Sigma}_j \right|} \right)$. 
\end{proposition}
Note that parameters $(\boldsymbol{\mu}_j,\mathbf{\Sigma}_j)$ for class $\mathcal C_j$ and $(\boldsymbol{\mu}_k,\mathbf{\Sigma}_k)$ for class $\mathcal C_k$ have been learned on the training dataset using the previously derived estimators.

\begin{proof}
The key idea of the proof is that the log-likelihood depends on $k$ only through the term $$\frac{1}{m} \log\left(\left|\mathbf{\Sigma}_k \right|\right) + \log\left((\mathbf{x}_i-\boldsymbol{\mu}_k)^T \mathbf{\Sigma}_k^{-1}(\mathbf{x}_i-\boldsymbol{\mu}_k)\right).$$
\end{proof}

\begin{remark}
This decision rule is close to the robust version of classic QDA except that we compare the log of the squared Mahalanobis distances rather than the squared Mahalanobis distances. Also, it is insensitive to the scale of $\mathbf{\Sigma}$.
\end{remark}

\section{Experiments on synthetic data}
\label{sec:3}
The proposed FEMDA method is compared to the following methods: classic QDA for Gaussian distributions, using classic and robust $M$-estimators (Robust QDA, see \cite{8} for details), QDA for $t$-distributions (t-QDA) \cite{20}, GQDA and RGQDA \cite{22}.

\textbf{Simulation settings:} Means of clusters are drawn randomly on the unit $m$-sphere and covariance matrices are generated with a random orthogonal matrix and random eigenvalues. The set up for the simulations is $m = 10$, $K = 5$, $N_{train} = 5000$, $N_{test} = 20000$ and $\tau \sim \mathcal{U}(1, m)$.


\textbf{Considered scenarios:} Points are drawn from four different families of ES distributions.

\begin{center}
\begin{tabular}{|l|l|}
  \hline
  Distribution family & Stochastic representation \\
  \hline
  
  generalized Gaussian & $\boldsymbol{\mu} + \Gamma(\frac{m}{2 \beta}, 2)^{\frac{1}{2 \beta}} \mathbf{\Sigma}^{\frac{1}{2}} \mathcal{U} \left( \mathcal{S}(0,1) \right)$ \\
  $t$-distribution & $\boldsymbol{\mu} + \mathcal{N}(0, \mathbf{\Sigma}) \sqrt{\frac{1}{\Gamma(\frac{\nu}{2}, \frac{2}{\nu})}}$ \\
  $k$-distribution & $\boldsymbol{\mu} + \mathcal{N}(0, \mathbf{\Sigma}) \sqrt{\Gamma(\nu, 1/\nu)}$ \\
      
  \hline
\end{tabular}
\end{center}

$\mathcal{U} \left( \mathcal{S}(0,1) \right)$ stands for the uniform distribution on the unit $m$-sphere. Shape parameter $\beta$ (resp. $\nu$) is drawn uniformly in $[0.25, 10]$ (resp. $[1, 10]$) for generalized Gaussian (resp. for $t$-distributions and $k$-distributions). 

Data generation scenario are identified as follows: $0.5GG - 0.3T - 0.3K$ corresponds to 50\% of the points for each cluster is drawn from a generalized Gaussian distribution, 30\% from a $t$-distribution and 20\%  from a $k$-distribution. 

Concerning the parameters, we use the following color code: \color{mygreen}$0.5GG - 0.3T - 0.2K$ \color{black}:  same $\beta$ and $\nu$ are used for all points across the same cluster and \color{myred} $0.5GG - 0.3T - 0.2K$ \color{black}: one different parameter is used for each point of each cluster.

While t-QDA and FEMDA rely on their own estimators, we will use either the classic empirical estimators (QDA and GQDA), or robust $M$-estimators (Robust QDA and RGQDA).



\begin{figure}[H]
\centering
\subfigure[Estimators\label{fig:1a}]{\includegraphics[width=4.2cm]{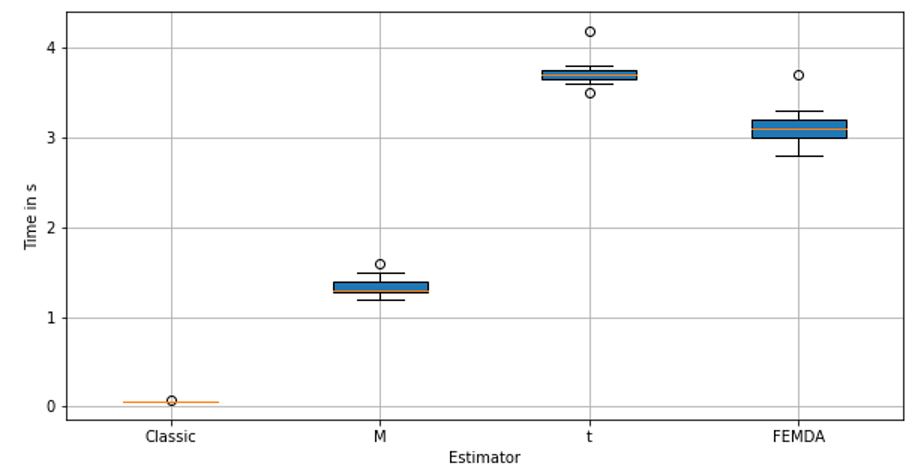}}
\subfigure[Decision rules\label{fig:1b}]{\includegraphics[width=4.2cm]{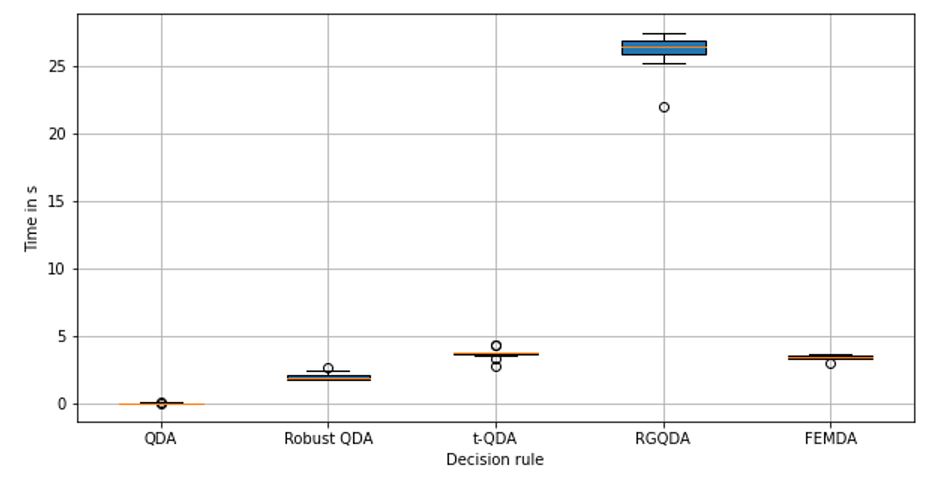}}
\caption{Convergence speed}
\label{fig:1} 
\end{figure}

As expected, one can see on Fig.\ref{fig:1a} that the classic empirical estimator is the fastest to be computed while $t$-estimator is slower because it requires the estimation of more parameters at each step. For t-QDA though, since the estimation of the degree of freedom is already optimized, the relative time gain will be smaller, making FEMDA and the other methods even faster than t-QDA. On Fig.\ref{fig:1b}, one observes that the speed for each decision rule is basically the convergence speed of the estimators used to compute the likelihood, except for GQDA that requires the estimation of an extra parameter.

\textbf{Results for the classification}

For several scenarios, specified in first column, Table \ref{table:1} displays the difference of accuracy between the obtained accuracy and the accuracy of the best method on the corresponding scenario : 
\vspace*{-.5cm}\begin{center}
\begin{table}[!h]
\begin{tabular}{|l|l|l|l|l|}
  \hline
  Scenario & QDA & t-QDA & GQDA & FEMDA \\

  \hline

  GG - T - K &  \cellcolor{black} & \cellcolor{black} & \cellcolor{black} & \cellcolor{black} \\
      
  \hline
    
    \color{mygreen} $1-0-0$ \color{black} & $-0.51$ & \cellcolor{myblue} $\textbf{76.27}$ & \cellcolor{mypowderblue} $-0.47$ & \cellcolor{myskyblue} $-0.02$ \\
      
  \hline
  
    \color{mygreen} $0-1-0$ \color{black} & \cellcolor{mypowderblue} $-0.64$ & \cellcolor{myblue} $\textbf{76.74}$ & $-0.69$ & \cellcolor{myskyblue} $-0.16$ \\
      
  \hline
    
    \color{mygreen} $0-0-1$ \color{black} & \cellcolor{mypowderblue} $-0.89$ & \cellcolor{myblue} $\textbf{76.43}$ & $-0.91$ & \cellcolor{myskyblue} $-0.12$ \\
      
  \hline
    
    \color{myred} $1-0-0$ \color{black} & $-0.59$ & \cellcolor{myblue} $\textbf{76.39}$ & \cellcolor{mypowderblue} $-0.58$ & \cellcolor{myskyblue} $-0.10$ \\
      
  \hline
  
    \color{myred} $0-1-0$ \color{black} & \cellcolor{mypowderblue} $-1.24$ & \cellcolor{myblue} $\textbf{77.08}$ & $-1.27$ & \cellcolor{myskyblue} $-0.21$ \\
      
  \hline
    
    \color{myred} $0-0-1$ \color{black} & \cellcolor{mypowderblue} $-1.08$ & \cellcolor{myblue} $\textbf{77.12}$ & $-1.17$ & \cellcolor{myskyblue} $-0.17$ \\
      
  \hline

    \color{mygreen} $\frac{1}{2}-\frac{1}{2}-0$ \color{black} & $-1.17$ & \cellcolor{myblue} $\textbf{80.85}$ & \cellcolor{mypowderblue} $-1.13$ & \cellcolor{myskyblue} $-0.39$ \\
    
  \hline
    
    \color{myred} $\frac{1}{2}-\frac{1}{2}-0$ \color{black} & $-1.31$ & \cellcolor{myskyblue} $-0.02$ & \cellcolor{mypowderblue} $-0.87$ & \cellcolor{myblue} $\textbf{80.59}$ \\
      
  \hline

    \color{mygreen} $\frac{1}{3}-\frac{1}{3}-\frac{1}{3}$ \color{black} & $-1.84$ & \cellcolor{myblue} $\textbf{80.79}$ & \cellcolor{mypowderblue} $-1.62$ & \cellcolor{myskyblue} $-0.04$ \\
    
  \hline
       
    \color{myred} $\frac{1}{3}-\frac{1}{3}-\frac{1}{3}$ \color{black} & $-2.17$ & \cellcolor{myblue} $\textbf{79.75}$ & \cellcolor{mypowderblue} $-1.75$ & \cellcolor{myskyblue} $-0.15$ \\
    
  \hline

\end{tabular}
\caption{Classification accuracy}
\label{table:1}
\end{table}
\end{center}
\vspace*{-1cm} In Table \ref{table:1}, one can see that GQDA performs better than QDA in scenarios with mixtures of distributions and evenly when only one type of distribution is used. However, GQDA performance does not compete with t-QDA and FEMDA, t-QDA being in most scenarios the best method but with a very slight improvement over FEMDA. This is due to the estimation of an extra parameter for t-QDA, namely $\nu$. The couterpart is that tQDA is slower than FEMDA.  
In table \ref{table:2}, We add some contaminated data using the distribution $\mathcal{N}(0, \Sigma_{noise})$ to simulated a contaminated point. One can see that FEMDA is the most robust to contamination. At a 25\% contamination rate, t-QDA is outperformed in almost all scenarios. Indeed, there are more parameters to estimate, and thus t-QDA is more sensitive to the contamination.

\vspace*{-.5cm}\begin{center}
\begin{table}[!h]
\begin{tabular}{|l|l|l|l|l|}
  \hline
  Scenario & t-QDA & FEMDA & t-QDA & FEMDA \\
  
   \hline
  Contamination & \multicolumn{2}{l|}{10\%} & \multicolumn{2}{l|}{25\%} \\
  
   \hline

  GG - T - K &  \cellcolor{black} & \cellcolor{black} & \cellcolor{black} & \cellcolor{black} \\
      
  \hline
    
    \color{mygreen} $1-0-0$ \color{black} & \cellcolor{mypowderblue} $-0.13$ & \cellcolor{myblue} $\textbf{70.41}$ & \cellcolor{mypowderblue} $-0.57$ & \cellcolor{myblue} $\textbf{61.25}$ \\
      
  \hline
  
    \color{mygreen} $0-1-0$ \color{black} & \cellcolor{myblue} $\textbf{71.70}$ & \cellcolor{mypowderblue} $-0.47$ & \cellcolor{mypowderblue} $-0.35$ & \cellcolor{myblue} $\textbf{62.00}$ \\
      
  \hline
    
    \color{mygreen} $0-0-1$ \color{black} & \cellcolor{myblue} $\textbf{70.80}$ & \cellcolor{mypowderblue} $-0.02$ & \cellcolor{mypowderblue} $-0.08$ & \cellcolor{myblue} $\textbf{61.47}$ \\
      
  \hline
    
    \color{myred} $1-0-0$ \color{black} & \cellcolor{mypowderblue} $-0.07$ & \cellcolor{myblue} $\textbf{70.03}$ & \cellcolor{mypowderblue} $-0.42$ & \cellcolor{myblue} $\textbf{61.29}$ \\
      
  \hline
  
    \color{myred} $0-1-0$ \color{black} & \cellcolor{myblue} $\textbf{70.98}$ & \cellcolor{mypowderblue} $-0.06$ & \cellcolor{mypowderblue} $-0.11$ & \cellcolor{myblue} $\textbf{61.51}$ \\
      
  \hline
    
    \color{myred} $0-0-1$ \color{black} & \cellcolor{myblue} $\textbf{71.01}$ & \cellcolor{mypowderblue} $-0.02$ & \cellcolor{mypowderblue} $-0.25$ & \cellcolor{myblue} $\textbf{61.52}$ \\
      
  \hline

    \color{mygreen} $\frac{1}{2}-\frac{1}{2}-0$ \color{black} & \cellcolor{myblue} $\textbf{75.53}$ & \cellcolor{mypowderblue} $-0.60$ & \cellcolor{myblue} $\textbf{65.43}$ & \cellcolor{mypowderblue} $-0.43$ \\
    
  \hline
    
    \color{myred} $\frac{1}{2}-\frac{1}{2}-0$ \color{black} & \cellcolor{myblue} $\textbf{74.72}$ & \cellcolor{mypowderblue} $-0.13$ & \cellcolor{mypowderblue} $-0.17$ & \cellcolor{myblue} $\textbf{64.55}$ \\
      
  \hline
        
    \color{mygreen} $\frac{1}{3}-\frac{1}{3}-\frac{1}{3}$ \color{black} & \cellcolor{myblue} $\textbf{74.09}$ & \cellcolor{mypowderblue} $-1.04$ & \cellcolor{mypowderblue} $-0.02$ & \cellcolor{myblue} $\textbf{64.42}$ \\
    
  \hline
       
    \color{myred} $\frac{1}{3}-\frac{1}{3}-\frac{1}{3}$ \color{black} & \cellcolor{myblue} $\textbf{73.44}$ & \cellcolor{mypowderblue} $-0.06$ & \cellcolor{mypowderblue} $-0.09$ & \cellcolor{myblue} $\textbf{63.45}$ \\
    
  \hline

\end{tabular}
\caption{Classification accuracy for contaminated data}
\label{table:2}
\end{table}
\end{center}

\vspace*{-1cm}\section{Results on real datasets}
\label{sec:4}
\subsection{Description of the datasets}

In this section, we present results on real datasets obtained from the UCI machine learning repository \cite{29}: \textbf{Spambase} where the objective is to classify emails between spams and non-spams. Attributes contain the frequency of use of usual words or characters; \textbf{Ecoli} where one wants to predict the localization site of a protein among 8 possible using 7 attributes about the cell that contains the protein; \textbf{Statlog Landsat Satellite} that contains multi-spectral values of pixels in 3*3 neighbourhoods in a satellite image. The goal is to predict the type of soil represented by the central pixel.

\subsection{Classification accuracy results}

The results have been averages over 100 simulations, and every 10, we reshuffle a new train and test set.

\begin{figure}[!h]
\centering
\subfigure[Spambase\label{fig:2a}]{\includegraphics[scale=0.17]{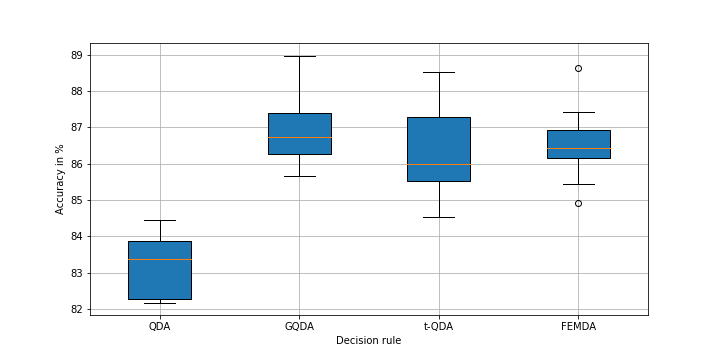}}
\subfigure[Ecoli\label{fig:2b}]{\includegraphics[scale=0.23]{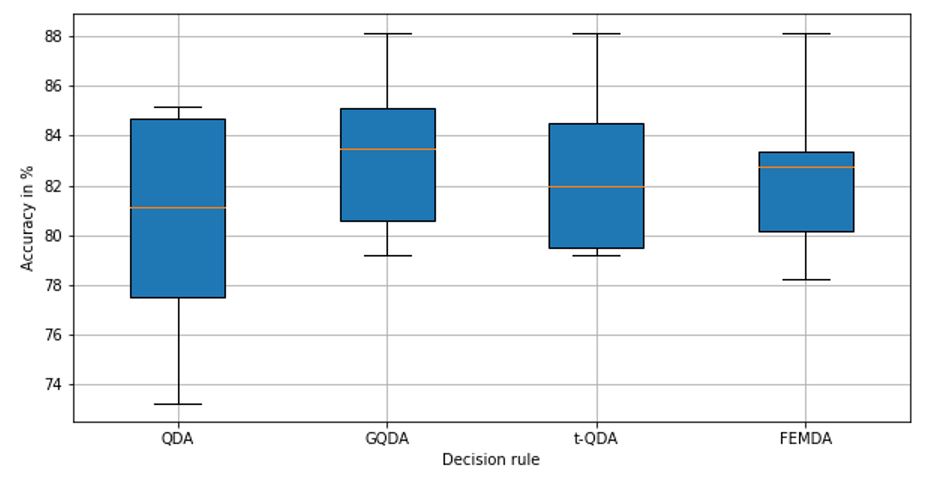}}
\subfigure[Statlog Landsat Satellite\label{fig:2c}]{\includegraphics[scale=0.23]{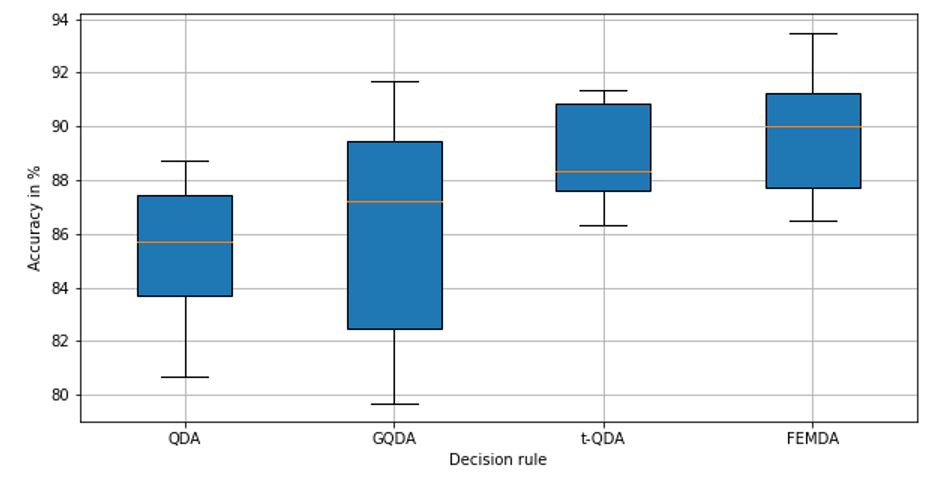}}
\caption{Median accuracy}
\label{fig:2} 
\end{figure}

We can see on Fig. \ref{fig:2a} and Fig. \ref{fig:2b} that for the Spambase and Ecoli dataset, GQDA slightly outperforms the other methods. FEMDA is better than t-QDA that also suffers from higher variance. It is worth noting that for those two datasets, GQDA is outperformed by LDA which shows that its good performances come from the ability to neglect the covariances if needed. Fig. \ref{fig:2c}  display the results obtained on the Statlog dataset. Again, GQDA slightly outperforms QDA but has much more variance. The two best methods are t-QDA and FEMDA with smaller variance.

\subsection{Performance under contaminated model}
\begin{figure}[!h]
\centering
\subfigure[Spambase\label{fig:3a}]{\includegraphics[scale=0.6]{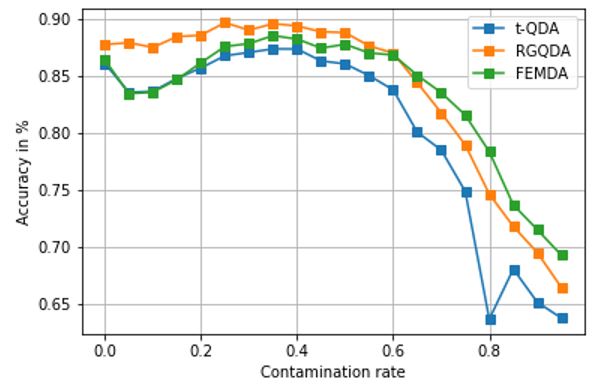}}
\subfigure[Ecoli\label{fig:3b}]{\includegraphics[scale=0.6]{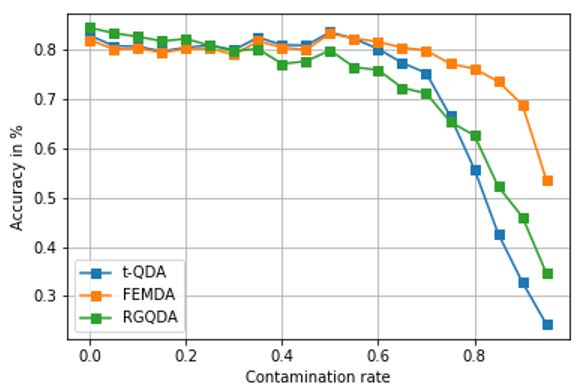}}
\subfigure[Statlog Landsat Satellite\label{fig:3c}]{\includegraphics[scale=0.6]{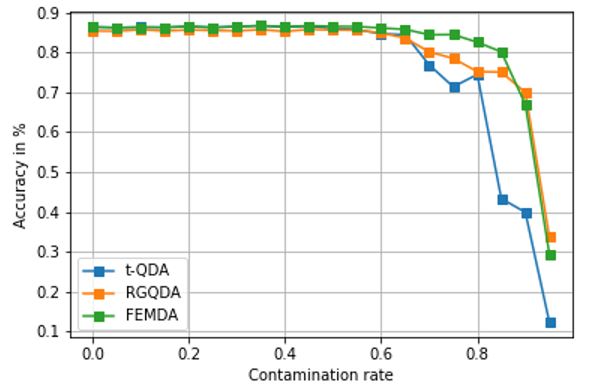}}
\caption{Contaminated data: for Spambase using $\mathcal{U}([0, 100]^{57})$; for Ecoli using $\mathcal{U}([0, 1]^{7})$ and for Statlog Landsat Satellite using $\mathcal{U}([0, 200]^{36})$}
\label{fig:3} 
\end{figure}
As detailed on Fig.\ref{fig:3}, the amplitude of the contaminated data changes from one dataset to another. We observe that even when the contamination rate is very high, we still observe good results. This can be explained by the two following reasons:
\begin{itemize}
    \item Most dataset used have well separated clusters : even a linear classifier (LDA) achieves very good performance.
    \item Contamination is mild, it is a random noise with no structure that could lead the classifier to consider all the noisy data as another cluster. These noisy points are well-handled by robust estimators thanks to the weighting.
\end{itemize}
On Fig.\ref{fig:3a} we can see that for the Spambase dataset, FEMDA starts to overwhelm GQDA at a 60\% contamination rate, and t-QDA at a 20\% contamination rate. The proposed method has less parameters to be estimated, and thus, it is less sensitive to noise and more robust. Concerning the Ecoli dataset, on Fig.\ref{fig:3b}, methods are not very impacted for low contamination rates. FEMDA and t-QDA remain very close. At a 50\% contamination rate, FEMDA becomes to outperform both t-QDA and RGQDA. FEMDA manages to preserve its performances up to a 70\% contamination rate, versus 50\% for other methods. Again, t-QDA is the most sensitive method to outliers and FEMDA is the most robust, being able to deal with much higher contamination rates. On the last dataset, Fig.\ref{fig:3c}, all methods obtain very similar results up to a 60\% contamination rate. t-QDA is the less robust and its performances start to erode quickly. FEMDA manages to uphold its performances up to a 80\% contamination rate, being again the most robust method to noise.

\section{Conclusion}
\label{sec:5}
In this paper, we presented a new highly robust discriminant analysis method that outperforms several state of the art methods for both simulated and real datasets. In this new approach, clusters no longer share the same covariance matrix, but only the same shape matrix. Sacrificing the scale of the covariance matrix allows us to gain flexibility in order to deal with non identically distributed observations. Moreover, the flexibility of such approach makes it particularly suitable to deal with heavy-tailed and contaminated data. Tests performed on simulated data show that our new approach has a computational speed comparable to t-QDA or QDA with plug-in robust estimators. Performances are almost as good as the best methods with clean data in various scenarios. When data are contaminated, the proposed FEMDA outperforms other robust methods in most scenarios. Simulations on real data also lead to the same conclusions. FEMDA performs as well as other methods in the presence of clean data and shows remarkable robustness when data is contaminated. It has the highest resilience to contamination. It can be seen as an enhancement of t-QDA: almost as good accuracy results but faster and much more robust.

\bibliographystyle{IEEEtran}
\bibliography{IEEEabrv,IEEEbib}
\end{document}